\documentclass[conference]{IEEEtran}

\usepackage{amsfonts}
\usepackage{url}
\usepackage{booktabs}
\usepackage{caption}

\usepackage{setspace}
\newcommand{\suchthat}{\mathrel{\mathop\supset}\kern-4.0pt$-$\kern-1.0pt$-~$}

\usepackage{multirow}
\usepackage{hhline}
\usepackage{amsmath}
\usepackage{amssymb}
\usepackage{array}
\usepackage{fixltx2e}
\usepackage{color}
\usepackage{mathtools,nccmath}
\usepackage{mathrsfs}
\usepackage{enumitem}

\DeclareMathOperator*{\argmin}{argmin}

\usepackage{amsmath,amssymb,amsthm}
\newtheorem{remark}{Remark}

\usepackage[latin9]{inputenc}

\ifCLASSINFOpdf
  \usepackage[pdftex]{graphicx}
\else  
\fi

\usepackage{fixltx2e}

\usepackage{epstopdf}
\usepackage{comment}
\usepackage{setspace}

\begin{document}

\title{Federated Learning for UAV Swarms Under Class Imbalance and Power Consumption Constraints}
\author{Ilyes~Mrad$^{\ddagger}$,~Lutfi~Samara$^{\ddagger}$,~Alaa~Awad~Abdellatif$^{\ddagger}$,~Abubakr ~Al-Abbasi$^{||}$,~Ridha~Hamila$^{\ddagger}$,~Aiman~Erbad$^{\dagger}$\\
$^{\ddagger}$Qatar University, Doha, Qatar, $^{||}$Qualcomm Inc., San Diego, USA \vspace{-0in}, $^{\dagger}$Hamad Bin Khalifa University, Doha, Qatar  \vspace{0mm}}

\maketitle  

\begin{abstract}
The usage of unmanned aerial vehicles (UAVs) in civil and military applications continues to increase due to the numerous advantages that they provide over conventional approaches. Despite the abundance of such advantages, it is imperative to investigate the performance of UAV utilization while considering their design limitations. This paper investigates the deployment of UAV swarms when each UAV carries a machine learning classification task. To avoid data exchange with ground-based processing nodes, a federated learning approach is adopted between a UAV leader and the swarm members to improve the local learning model while avoiding excessive air-to-ground and ground-to-air communications. Moreover, the proposed deployment framework considers the stringent energy constraints of UAVs and the problem of class imbalance, where we show that considering these design parameters significantly improves the performances of the UAV swarm in terms of classification accuracy, energy consumption and availability of UAVs when compared with several baseline algorithms.  
\end{abstract} 

\section{Introduction}

Unmanned aerial vehicles (UAVs) are currently being deployed to enhance services across a multitude of applications \cite{uavdesr}. Such applications cover for instance aerial surveillance in law enforcement applications, cargo transport and conducting reconnaissance tasks in military applications. Although the deployment of UAVs introduces a new design degree of freedom brought by the possibility of three dimensional UAV mobility, this brings up the issue of the stringent energy constraint due to the limited battery lifetime that operates the UAV. Such energy constraints motivates the investigation of their performances while carrying out such vital tasks.

Federated Learning (FL) is an emerging solution that promotes decentralized model training \cite{konevcny2016federated}. FL can be exploited as it does not require the conventional approach of exchanging local training data between the client and the server. Instead, the use of FL emerges as a potential candidate to distribute machine learning tasks instead of relying on a centralized processing node, for example a ground processing node. In such setups, each UAV trains its learning model based on its own collected data, and then exchanges the \textit{learned parameters} with other UAVs in the swarm to reach a global consensus. FL can be utilized by UAVs to deliver various vital tasks such as target recognition or path planning. In the literature, various recent works have considered the problem of incorporating FL into UAV swarms \cite{zeng2020federated}, where a UAV leader aggregates the local learning-related parameters and updates the global ones over several rounds of communications within the swarm members.  

As the spatial dimension is significantly larger in an aerial networks environment in contrast to its ground-based counterparts, FL can minimize the communications between each UAV and ground-based base-stations, where communicating with ground BS nodes is made exclusive to the swarm leader. 

Recently, FL has been considered as a distributed Machine Learning (ML) approach to reach an accurate global learning model while considering several design constraints. The efficiency of FL was assessed in \cite{synchFL} through extensive experiments on various datasets. This work considered the typical FL model, where a group of users exchange their updated learning  models with a central node responsible for averaging and constructing a global model with a fixed global aggregation frequency.  
Then, several studies have extended this model. For instance, the authors in \cite{zhao_federated_2018} investigated the effect of non-Independent and Identically Distributed (non-IID) data on the performance of FL. In particular, the work in \cite{zhao_federated_2018} has shown that the non identical distribution of the data at different users can significantly decrease the obtained accuracy. Convergence analysis of Federated Averaging (\textsc{FedAvg}) algorithm is presented in \cite{Convergence_FedAvg} for convex and smooth problems while considering non-IID data. In \cite{Wang2019}, the authors presented an adaptive FL aggregation scheme that targets minimizing the learning loss under resource-constrained environment, while considering the non-IID data problem.   
In \cite{flOptimizationModel}, a modified FL algorithm is proposed using a local surrogate function that allows each participant to update its local model up to a certain accuracy level. Moreover, a resource allocation optimization problem is solved to address the trade-off between the training time of the proposed algorithm and participant's energy consumption.  In \cite{Naram2020}, the effect of the non-IID data in edge-assisted FL environment is investigated, while assessing the main parameters that affect the learning performance.

The problem of class imbalance arises in learning problems when the training data corresponding to majority classes accounts for a larger portion of the overall training data, while the ones corresponding to minority classes account for a lower portion. This may result in a reduction in classification accuracy of minority classes, as investigated in \cite{li2019convergence}. In UAV networks, data collected by UAVs may experience the problem of class imbalance like in \cite{jenssen2019intelligent}, where it was encountered when UAVs were used to inspect power lines. 

Several recent articles addressed the problem of FL in UAV networks. For instance, in \cite{zeng2020federated}, the authors conducted a convergence analysis for FL while considering several design factors such as wireless propagation effects. In \cite{donevski2021federated}, the authors proposed a drone trajectory optimization approach to serve FL networks to address the problem of learning discrepancies between nodes. However, the ML task was not selected to be drone-mounted. In \cite{zhang2020federated}, image classification was performed by UAVs while carrying out an exploration task and minimizing the computational aspect on the centralized ground fusion center, and beamfroming is performed by each of the UAVs to enhance the classification accuracy at relatively low communication cost.

In contrast to the discussed literature, we propose a solution to the problem of class imbalance in FL for energy-constrained UAV networks, where the aim is to improve the classification accuracy while considering the limited availability of UAVs due to their stringent energy constraints. While the availability of UAVs is determined by the expected service time of a UAV given its energy level, the classification accuracy is enhanced by developing a selection algorithm, where the set of UAVs that delivers the lowest class imbalance will be selected. Simulation results show that the proposed algorithm provides significant classification accuracy and overall power consumption gains when compared with several baselines. Such results pave the way to explore the deployment of FL in UAV networks, especially with the emergence of novel computers that could be integrated to UAVs and deliver computationally challenging services such as image processing\footnote{https://www.dji.com/manifold-2}.

%


\section{System Model} \label{sysmodel}

\begin{figure}[htbp]
\centering
    \includegraphics[width=2.4in]{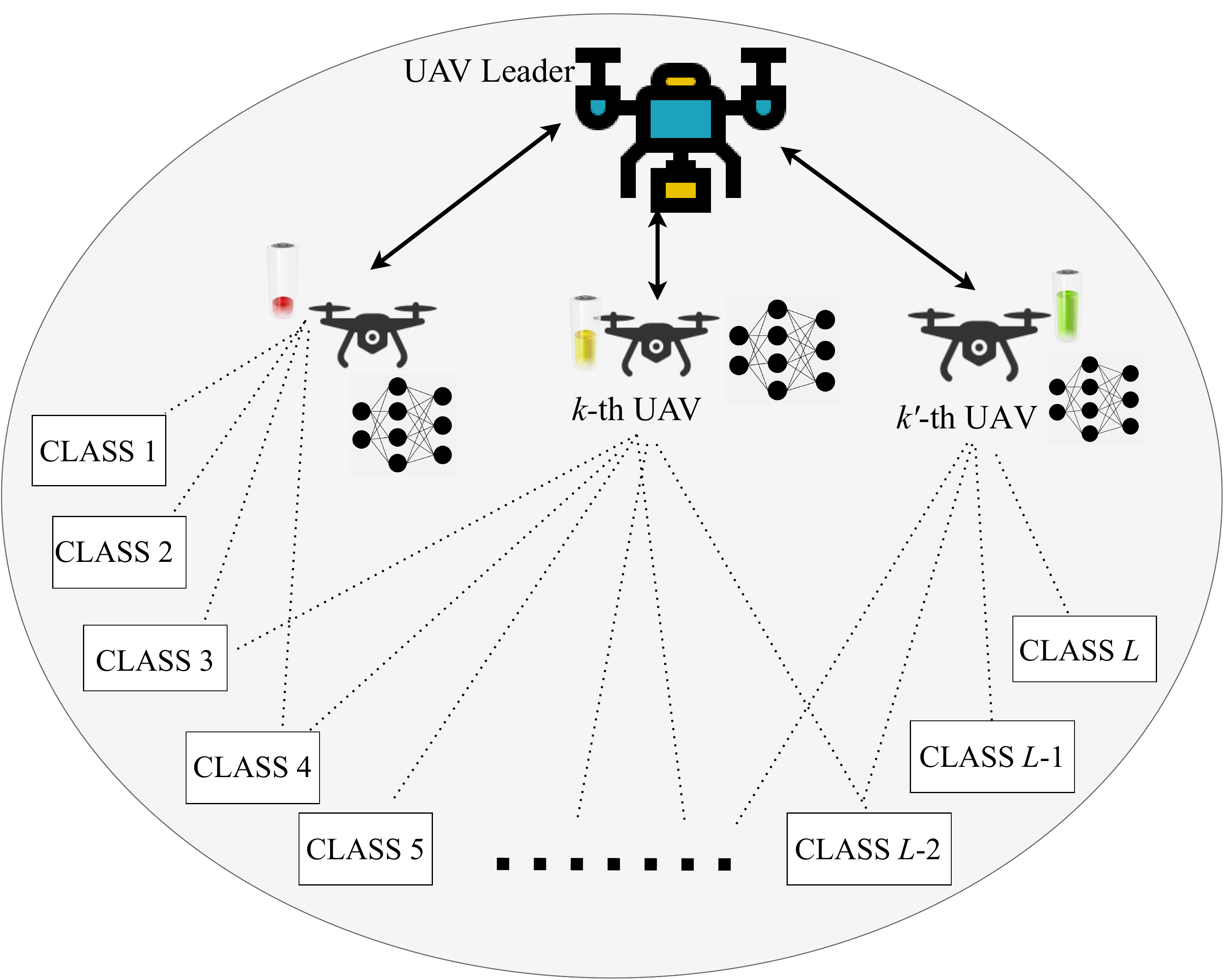}
    \caption{System model.}
    \label{fig:sm}
\end{figure}

We assume having a swarm of UAVs, where the $k^{th}$ UAV $k\in\{1 \hdots K\}$ collects a set of input data denoted by $\{x_{k1},x_{k2},\hdots,x_{k_{Q_k} }\}$, and $Q_k$ denotes the number of collected samples at the $k^{th}$ UAV. We assume that $x_{kq}$, $q\in\{1,\hdots,Q_k\}$ corresponds to a single output $y_{kq}$, and thus yielding an output vector $\{y_{k1}, y_{k2},\hdots,y_{kQ_k}\}$. The system model under consideration is depicted in Fig. \ref{fig:sm}, where each UAV is shown to be exposed to specific data, based on the considered application, such as target recognition and localization. The available UAVs will train their own models and then send their weights updates to a more capable UAV leader while aiming at reaching global convergence using FL. The weights update is done through a slotted wireless communication based user multiplexing scheme between the UAV leader and the UAV swarm members. We note that even when the UAV leader needs to recharge/swap its battery, its transceiver and processors may continue operating to deliver its FL based tasks while charging/swapping its battery. 

In this paper, such setup mimics the case of training an object recognition ML-based model, where a swarm of UAVs is deployed to learn a global recognition model using an FL approach. While training the ML model, some UAVs will be exposed to certain data classes more-so than others, hence creating the problem of class imbalance  \cite{jenssen2019intelligent,kellenberger2018detecting}.

Furthermore, as UAVs are governed with stringent energy constraints, it is mandatory to select the optimal subset of UAVs that will reliably complete the collective learning task. 
We assume that throughout a time window $\tau_t$, $M$ out of $K$ UAVs will be available due to the need for $K-M$ UAVs to recharge their empty batteries/return back to a UAV docking station at the next time window $\tau_{t+1}$. In that event, learning is not possible and the UAV will not contribute in FL. More details on the considered FL model and UAV power consumption model are provided in the next subsections.

\subsection{Federated Learning Model}

In FL, the convergence of the training process requires each learning vector at the UAV to reach $\mathbf{w}^*$, where the following problem is solved
\begin{align}
\argmin_{\mathbf{w}\in \mathbb{R}^d } \sum_{k=1}^K \frac{Q_k}{n}F_k(\mathbf{w}),
\end{align}
where $F_k(\mathbf{w})=\frac{1}{Q_k}\sum_{i\in \mathcal{S}_k} f(\mathbf{w},x_{ki},y_{ki})$ captures the overall loss function and $\mathcal{S}_k$ is the set of indexes for the $k^{th}$ user's data points, $n=\sum_{k=1}^K Q_k $. Moreover, the loss function $F(\mathbf{w})$ can take any form depending on the designer's choice or the application.
An iterative update technique can be adopted to solve the above problem, where the UAV leader generates an initial global model $\mathbf{w}_t$ that will be broadcasted to the UAV swarm, where it will be used over each of the UAVs' local data to then send the updated trained model back to the UAV leader. After receiving all updates from all UAV swarm members, the UAV leader will aggregate the received FL vectors, and will update it again to be re-sent to the UAV swarm members. This process will be repeated over time until the gap between the loss function $F(\mathbf{w})$ and $F(\mathbf{w}^*)$ gets below a threshold $\epsilon$, i.e. $|F(\mathbf{w})-F(\mathbf{w^*})| \leq \epsilon$. Each UAV will compute $\nabla F_k(\mathbf{w}_t)$, the average gradient related to its data, so then the UAV leader collects the computed gradients and applies 
\begin{align}
    &\mathbf{w}^k_{t+1} \leftarrow \mathbf{w}^k_t - \lambda  \nabla F_k(\mathbf{w}_t),\\
    &\mathbf{w}_{t+1} \leftarrow \sum^K_{k=1} \frac{Q_k}{n} \mathbf{w}^k_{t+1},
\end{align}
where $\lambda$ is the learning rate. Furthermore, the local update at the $k^{th}$ UAV can be computed several times by applying
\begin{align}
    \mathbf{w}^k \leftarrow \mathbf{w}^k - \lambda \nabla F_k(\mathbf{w}^k).
\end{align}
Then, model averaging at the UAV leader node can be implemented. This approach is termed (\textsc{FedAvg}). 
The \textsc{FedAvg} algorithm is governed by three main parameters: a) The portion of the number of available UAVs $M$, b) the number of training passes $E$ applied by each UAV on its local dataset for each communication round, c) the minibatch size $B$ used to update the model at the $k^{th}$ UAV, and d) the number of communication rounds between the UAV leader and the UAV swarm members to converge to the global learning model.

\subsection{ UAV Power Consumption Model }

Unlike ground-based nodes, UAVs are constrained by their energy demands making power consumption a necessary design consideration \cite{samara2021secure}. In this setup, we rotary-wing UAVs for their ease of deployment. As the propulsion power is the most significant part of the energy consumption model of a UAV \cite{zeng2017energy}, the communication and processing required powers are ignored. The propulsion power consumption is computed by applying
\begin{align}\nonumber
P_k &= P_0 \left( 1 + \frac{3V^2_k}{U^2_{tip}} \right) + P_i \left( {\sqrt{ 1 + \frac{V^4_k}{4v^4_{0}} } - \frac{V^2_k}{2v^2_{0}} }\right)^{\frac{1}{2}}\\
&  + \frac{1}{2} d_0 \rho s A V^3_k ,
\end{align} 
where $V_k$ is the speed of the $k^{th}$ flying UAV. Moreover, $P_0$ and $P_i$ are the blade profile power and the induced power during hovering, respectively, which can be written as
\begin{align} 
P_0 = \frac{\delta}{8} \rho s A \Omega^3 R^3,
\end{align}
and 
\begin{align}
P_i = (1+\iota) \frac{\mathcal{W}^{\frac{3}{2}}}{\sqrt{2\rho A}},
\end{align}
$U_{tip}$ is the tip speed of the rotor blade, $s$, $d_0$, $\rho$ and $A$ denotes the rotor solidity, the fuselage drag ratio, the air density and the rotor disc area, respectively, $v_{0}$ is the mean rotor induced velocity while hovering, $\mathcal{W}$ is the UAV weight in Newton, $R$ is the rotor radius, $\delta$ is the profile drag coefficient, $\iota$ is the incremental correction factor to the induced power and $\Omega$ is the blade angular velocity in radians per second. Further, the power consumption during the hovering status of the UAV is  
\begin{align}
P_h = P_0 + P_i.
\end{align}
In our formulation, the location of the UAV docking station must be known to all UAVs in order to calculate the expected power consumption figures.

\section{Problem Formulation and Proposed Solution}

In this section, we aim at developing a framework to perform UAV selection in order to guarantee a reliable and stable performance of FL while considering the problem of class imbalance. It is worth noting that the number of classes at each UAV must be exchanged between the swarm members and the UAV leader. The optimization problem is
\begin{subequations}
\begin{align} \label{eq:c2}
    \min_{\mathcal{A}  \in \Gamma}   & \sum_{k\in \mathcal{A}} P_k \\ 
     \textrm{s.t.} & \  \beta_i\geq \beta_{th}, \ i\in \{1,\hdots,N\}\label{con1} \\ 
     & \textrm{card}\left(\bigcup_{k\in \mathcal{A}} \mathcal{C}_k \right) = L, \label{con2}\\ 
     & \sum_{k'\in \mathcal{A'}} \sum_{l=1}^{L} \frac{Q_{k'}}{n} |\mathbf{C}_{k',l} - {O}_{k'} | \leq \varepsilon, \label{con3} 
\end{align}
\end{subequations}
where $\mathcal{C}_k$ represents the set of indexes of active classes for the $k^{th}$ UAV,  $C_{k,l}$ is the data size for the $k^{th}$ UAV at the $l^{th}$ class,  ${O}_{k} = \frac{1}{Q_k} \sum_{l=1}^L {C}_{k,l} \ \forall k\in \mathcal{A}$, $L$ is the number of classes that is specific to the learning problem and $\varepsilon$ is a threshold that can be set by the designer. Note also that
\begin{align}
    \textrm{card}(\mathcal{A'}) = \min \left( \textrm{card}\left(\mathcal{A}\right) \right) \ni \textrm{card}\left(\bigcup_{k\in \mathcal{A}} \mathcal{C}_k \right)   = L, \label{con4}
\end{align}    
where $\mathcal{A}^{'} \in \Gamma^{'}$, and $\Gamma^{'}$ contains all combinations that cover $L$ classes. The constraint in (\ref{con1}) guarantees that the selected UAVs will have battery levels that are greater than $\beta_{th}$, while constraint (\ref{con2}) insures that the set of selected UAVs will collectively cover data-sets belonging to all $L$ classes, and the constraint in (\ref{con3}) insures the uniformity of the distribution of the data-sets between the classes.

In other words, our aim is to maximize the availability of the UAVs so that the process of convergence to the global model will not be interrupted by the absence of a UAV(s). The availability of a UAV is determined first by setting a threshold on the UAV battery level, thus disregarding the presence of a UAV that has a low battery level as its presence is not guaranteed. Then, we select the set $\mathcal{A}^*$ of UAVs such that the total consumed power for all UAVs is minimized while guaranteeing that all classes are available.

\begin{remark}
The problem in (\ref{eq:c2}) may not be feasible to solve as the condition in (\ref{con2}) may not be reached. Hence, a feasibility problem can be formulated by replacing $L$ in (\ref{con2}) with $L'$, where $L'$ can be calculated by applying
\begin{align}
    L' = \max_{k\in \mathcal{A}} \left( \textrm{card}\left(\bigcup_{\mathcal{A}\in \Gamma} \bigcup_{k\in\mathcal{A}} \mathcal{C}_k \right) \right) ,
\end{align}
where $\Gamma$ is a set containing all possible combinations of the selected UAVs. Thus, $L$ in (\ref{con2}) will be replaced by $L'$ defined above.
\end{remark}

To solve the formulated optimization problem, we propose the following algorithms. $\textsc{Algorithm}$ 1 is aimed to satisfy the Eqs. (\ref{con2}), (\ref{con3}) and (\ref{con4}). Hence, we have considered that Eq. (\ref{con1}) is already satisfied, i.e. $\psi$ contains only the indexes of UAVs with battery levels greater that $\beta_{th}$. At first, we satisfy the constrains defined in Eqs. (\ref{con2}) and (\ref{con3}). Then, we removed the combinations that have a number of UAVs greater than $N_3$ which is the minimum of number of UAVs per combination. Hence, we are minimizing the total power consumption.



  

   \begin{figure}[htbp]
     \centering
     \includegraphics[trim={3.5cm 3cm 4.25cm 3cm},height=5in,width=3.1in]{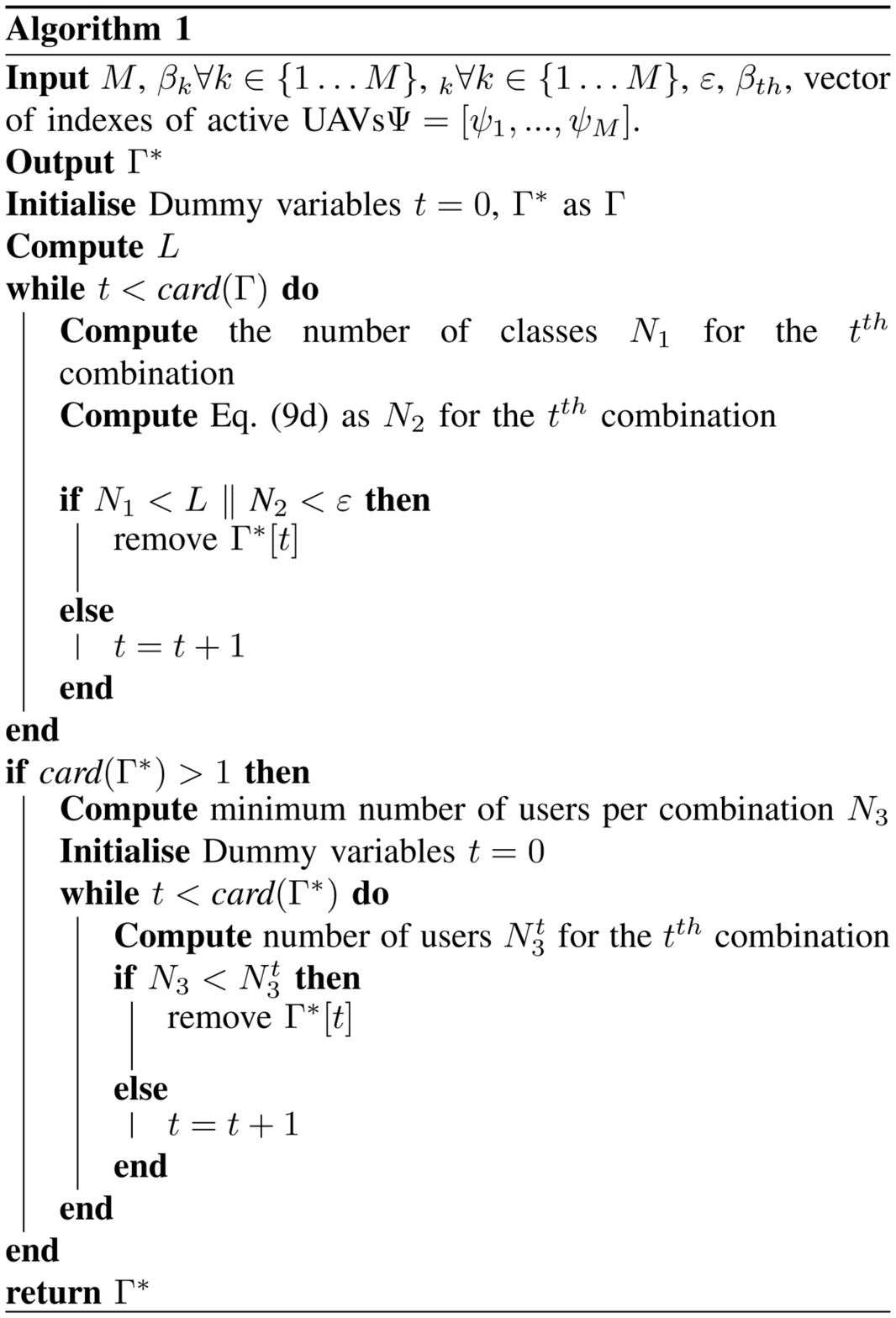}
   \end{figure}

As for \textsc{Algorithm} 2, it addresses the availability of UAVs by maximizing the minimum of the available UAV battery levels, and then maximizes the sum in case of not having a unique solution. The first two loops are aimed to find the combination that has the highest minimum battery level that guarantees the maximum availability of all UAVs of the combination. Moreover, if the problem still does not yield a unique solution, the two last loops will be aimed to return the combination that has the highest aggregation of battery levels.

\section{Numerical Results}

In this section, we present the numerical results, where we compare the proposed solution with several baseline algorithms. The implementation of the FL algorithm that was used throughout the paper was adopted from \cite{shaoxiong}. All UAVs are assumed to be assigned to a fixed location and hover, mimicking a surveillance scenario. The parameters related to the UAV power consumption model are given in Table \ref{table2}.


 


%


   \begin{figure}[htbp]
     \centering
     \includegraphics[trim={3.25cm 3cm 9cm 3cm},height=7in,width=3.1in]{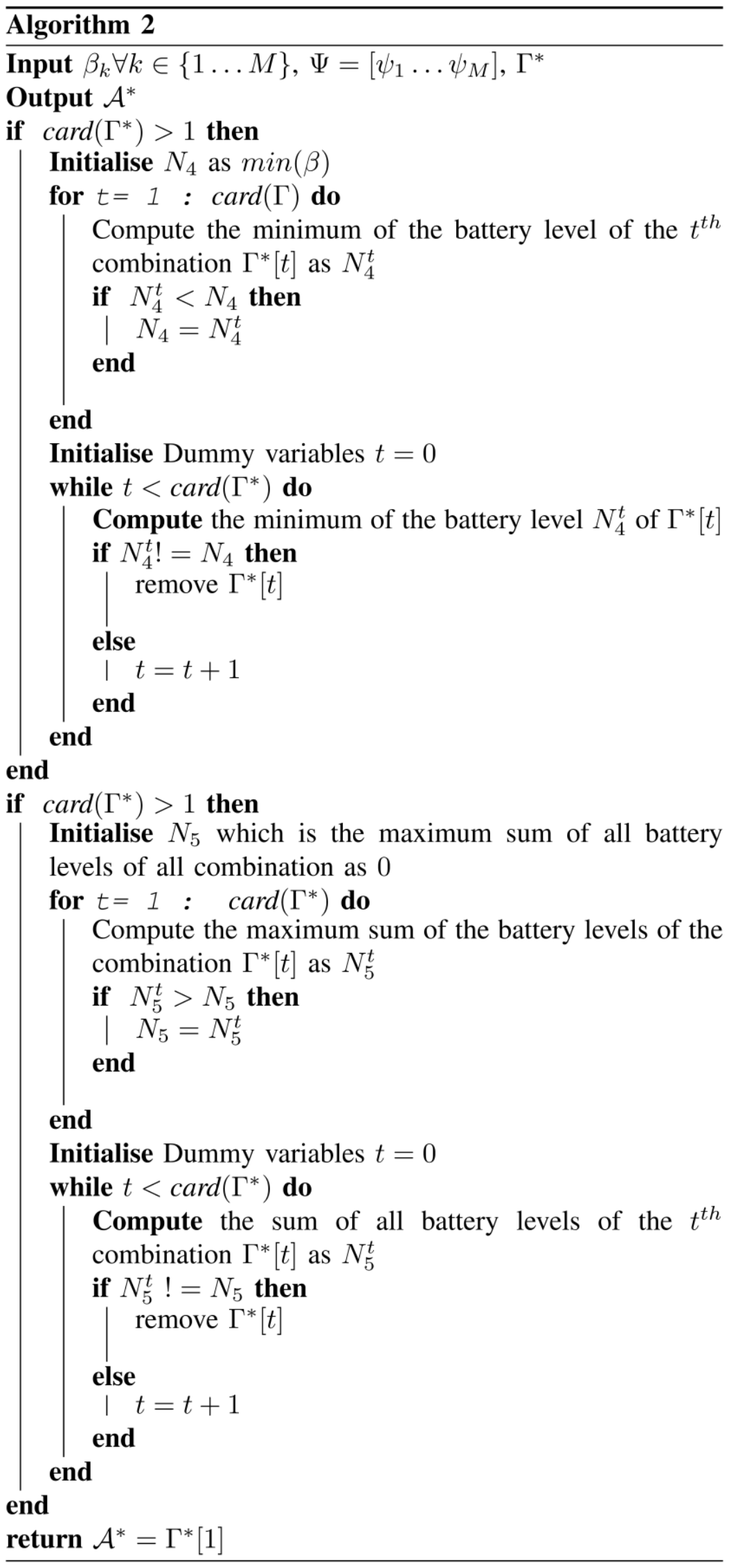}
   \end{figure}

\begin{table} 
\centering
\begin{tabular}{|c|c|c|c|}
\hline
 Parameter & Value & Parameter & Value       \\ \hline\hline
 $\mathcal{W}$       & $50$ Newtons   & $\rho$    & 1.225 kg/m$^3$\\ \hline
 $R$       & 0.25 m    & $A$       & 0.1963 m$^2$  \\ \hline
 $\Omega$  & 400 Radians/s   & $U_{tip}$    & 100   \\ \hline
 $d_0$     & 0.3    & $k$       & 0.1   \\ \hline
 $v_o$     & 10.2 m/s  & $\iota$     & 0.1 \\ \hline
 $\delta$  & 0.012   & $V$       & 18.46 m/s \\ \hline
 $s$       & 0.05    &  $K$  & 10 \\ \hline

\end{tabular}
\caption{Parameters used throughout the simulations.}\label{table2}
\end{table}
Moreover, the learning rate is $\lambda=0.01$, the number of training passes each UAV performs over its local data ($E$) is equal to 5, and the local minibatch size ${B}$ is equal to 10. Furthermore, the UAVs are assumed to be randomly distributed in a circle of radius 1 km with an altitude $\eta=100$ m. To test our proposed approach, the MNIST dataset is used, where we perform a digit recognition task, having a total of 10 classes. Note that this dataset was selected for the sole purpose of demonstrating the improvements made by adopting the proposed approach, and the type of the used dataset by itself is irrelevant at this stage. Moreover, we use a 2-hidden layer multilayer-perceptron neural network with 200 units, and each adopting ReLu activation functions (199210 total parameters). In the simulation setup, we define the class imbalance rate $\mu$ (unitless), which controls the number of active classes for a certain UAV. For instance, $\mu=0$ means that a UAV has data in all available classes, while if $\mu=0.9$ signifies that the UAV has a single visible class. This design parameter allows us to partially control the class imbalance at each UAV. Moreover, we assume that equal amounts of data are distributed amongst the available UAVs.

\begin{figure}
\centering
    \includegraphics[height=1.8in,width=3.1in]{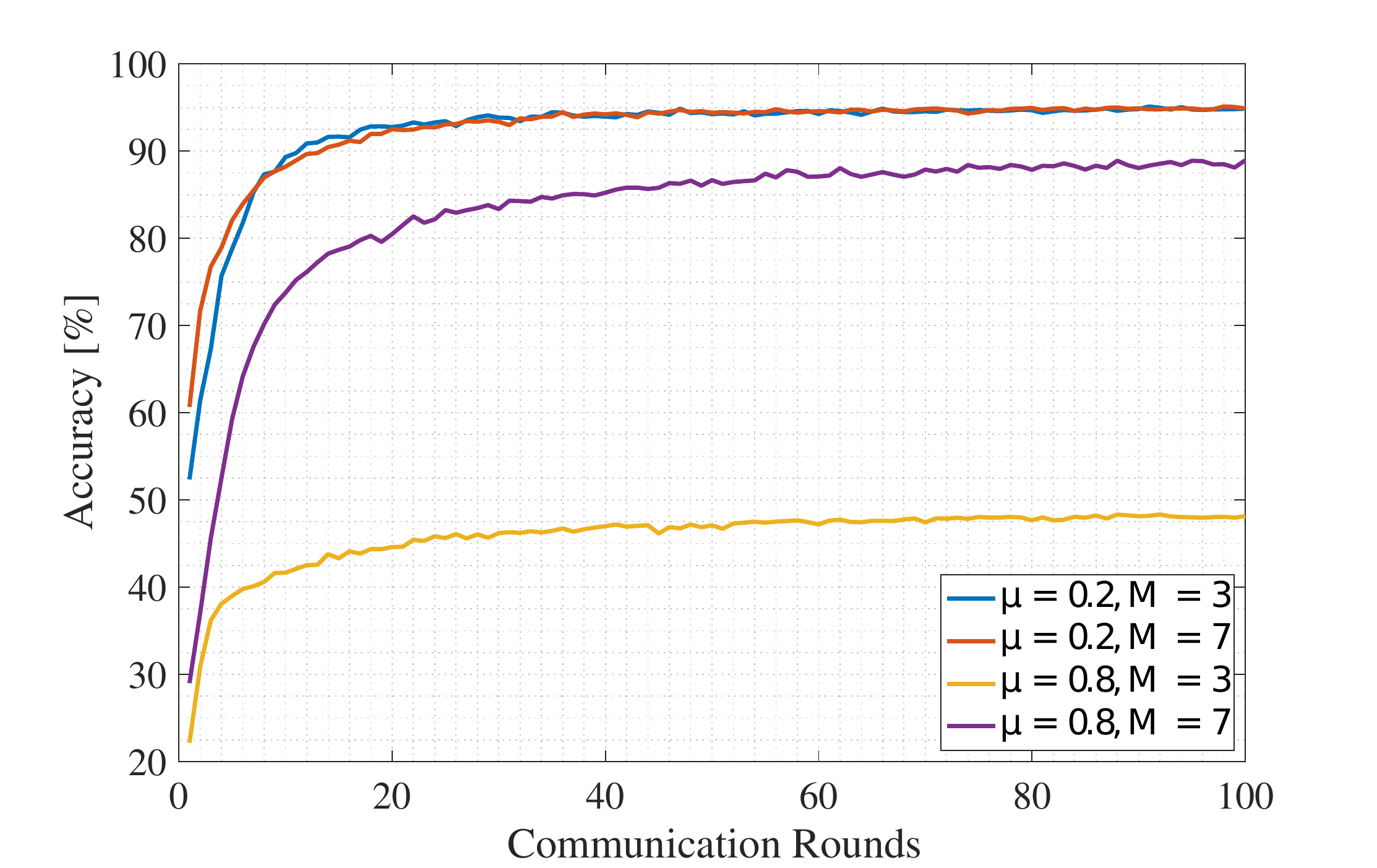}
    \caption{ Testing accuracy vs. comm. rounds.}
     \label{fig:f1}
\end{figure}

To visualize the effects of the number of available UAVs $M$ and the class imbalance rate on the testing accuracy, we begin by depicting the result in Fig. \ref{fig:f1}, where the testing accuracy is plotted vs. the number of communication rounds. As the number of communication rounds increases, the testing accuracy also increases, while saturating to different values depending on $M$ and $\mu$. When $\mu$ is high, for instance $\mu=0.8$, and a low number of UAVs ($M=3$) is selected, the testing accuracy is observed to be significantly low. For the same $\mu$, the testing accuracy is observed to improve significantly, reaching around $89\%$. It is also worth noting that when $\mu=0.2$, the number of active users $M$ has a relatively negligible effect when it comes to the testing accuracy, thus emphasizing on the importance of UAV selection and the design constraint in (\ref{con2}). 
\begin{figure}
\centering
    \includegraphics[height=1.8in,width=3.1in]{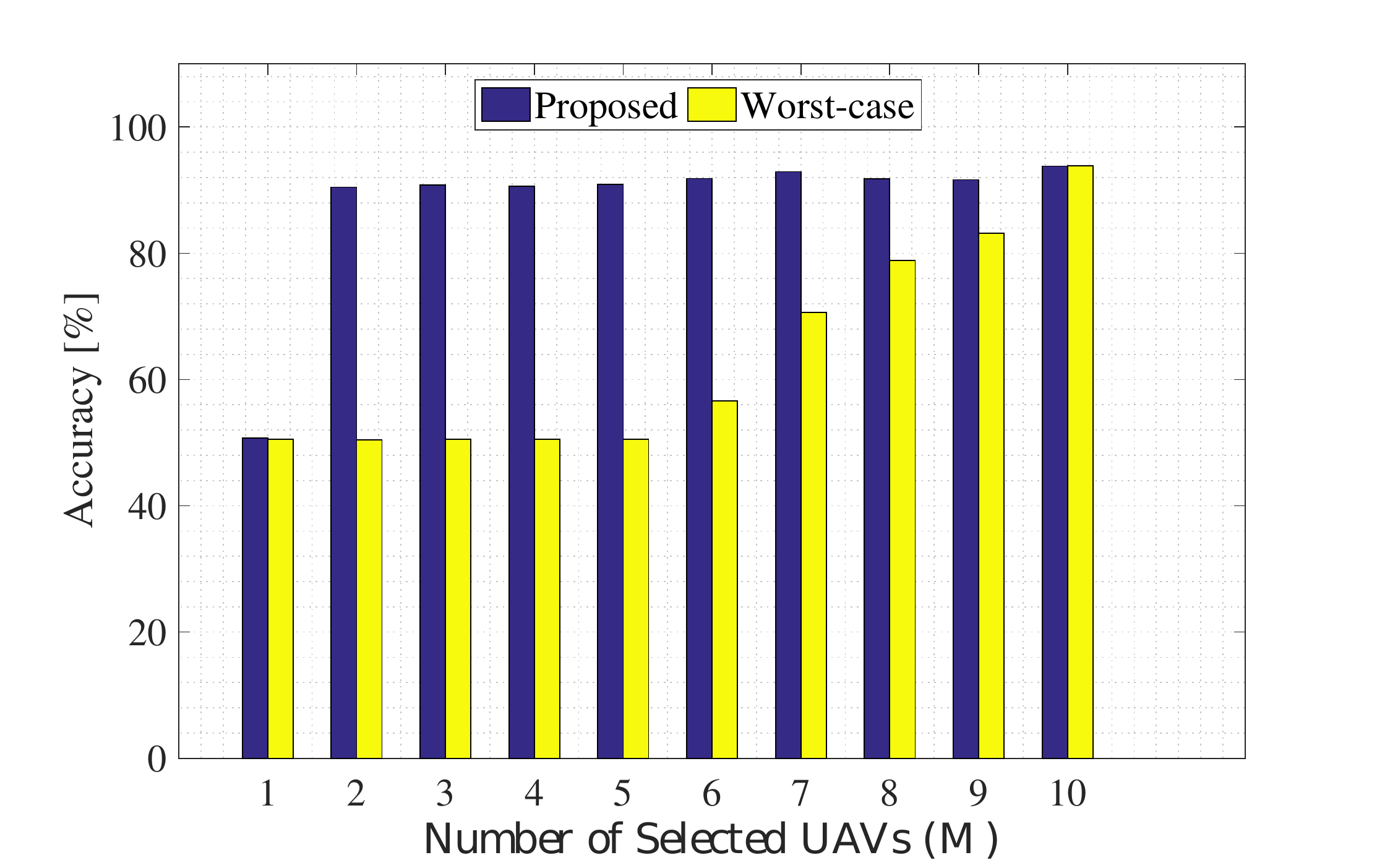}
    \caption{ Testing accuracy vs. $M$ of our proposed UAV selection scheme and the worst-case solution. $\mu=0.5$.}
    \label{fig:f2}
\end{figure} 
\begin{figure}
\centering
    \includegraphics[height=1.8in,width=3.1in]{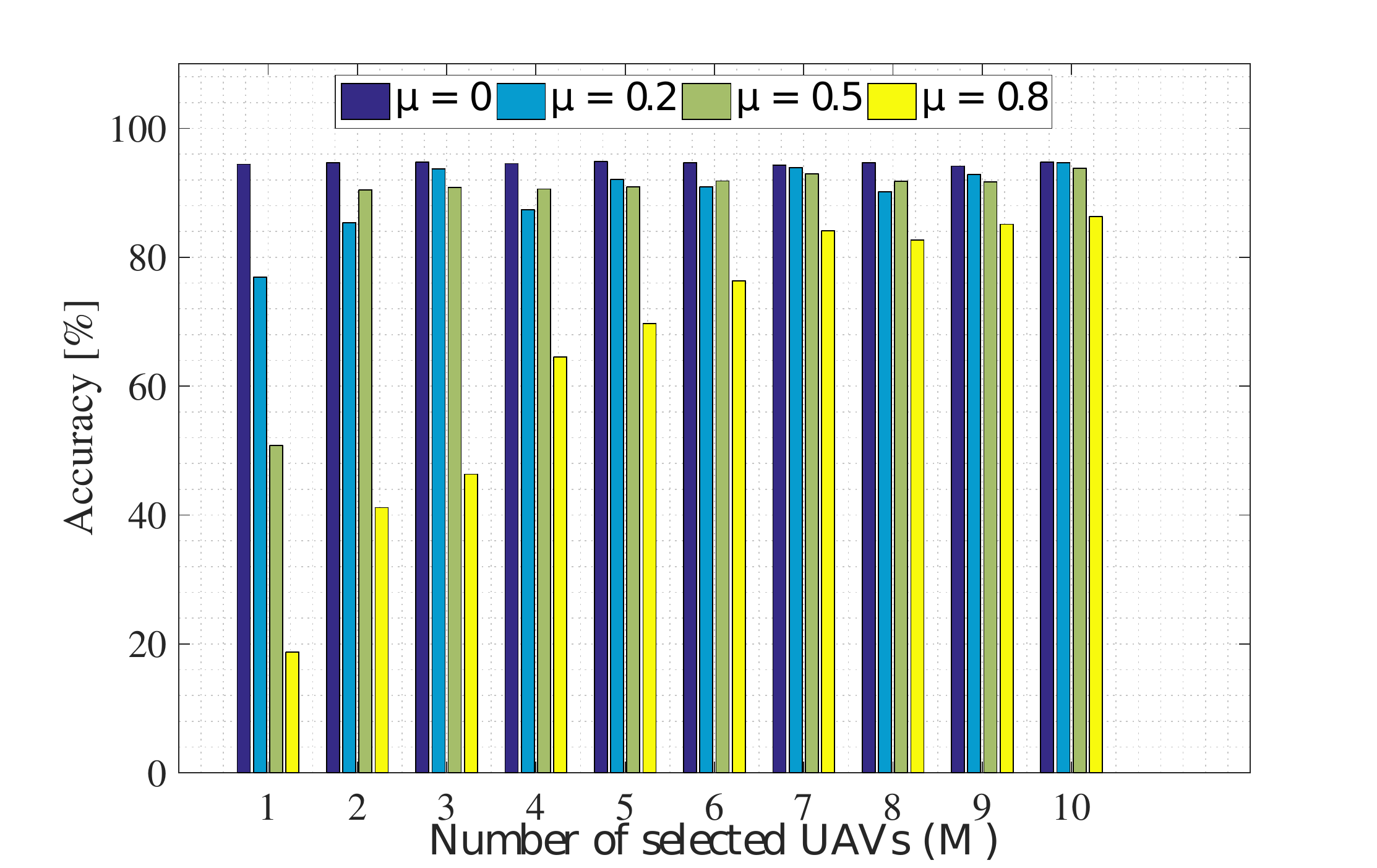}
    \caption{ Testing accuracy vs $M$, varying $\mu$.}
    \label{fig:f3}
\end{figure}
To highlight the improvements made by adopting the proposed algorithm, Fig. \ref{fig:f2} depicts the testing accuracy vs. the number of selected UAVs, where we plot the proposed approach explained in Section III when compared with a worst-case scenario baseline. In this setup, an equal number of selected UAVs was assumed and $\mu=0.5$. For the worst-case baseline, however, it was assumed that the selected UAVs do not conform with the second constraint in Eq. (\ref{eq:c2}), but selects UAVs that are exposed to similar classes. The difference in the testing accuracies is very well pronounced between the two approaches, hence highlighting the importance of UAV selection based on the available data classes. Moreover, the worst-case baseline is heavily dependent on the number of UAVs, while for our proposed approach, selecting 2 UAVs already improves the testing accuracy to around $90\%$.

Fig. \ref{fig:f3} shows the testing accuracy vs. the number of selected UAVs while varying $\mu$. For a fixed value of $\mu$, the accuracy is increased when augmenting the number of selected UAVs, except for the case when there is no class imbalance and $\mu=0$. However, we would like to note that the testing accuracy may not increase for all cases when $\mu$ is increased. For instance, when $M\in\{2,4,6,8\}$, the testing accuracy when $\mu=0.5$ is better than the case when $\mu=0.2$, which is counter intuitive since a lower $\mu$ should yield a better accuracy. This is due to the fact that although when $\mu=0.2$ means that each UAV has eight out of ten active classes, the eight active classes experience sever class imbalance, while when $\mu=0.5$, all active classes have equal amounts of data in them. Hence, this justifies the constraint in (\ref{con3}).

\begin{figure}[htbp]
     \centering
     \includegraphics[height=1.9in,width=3.1in]{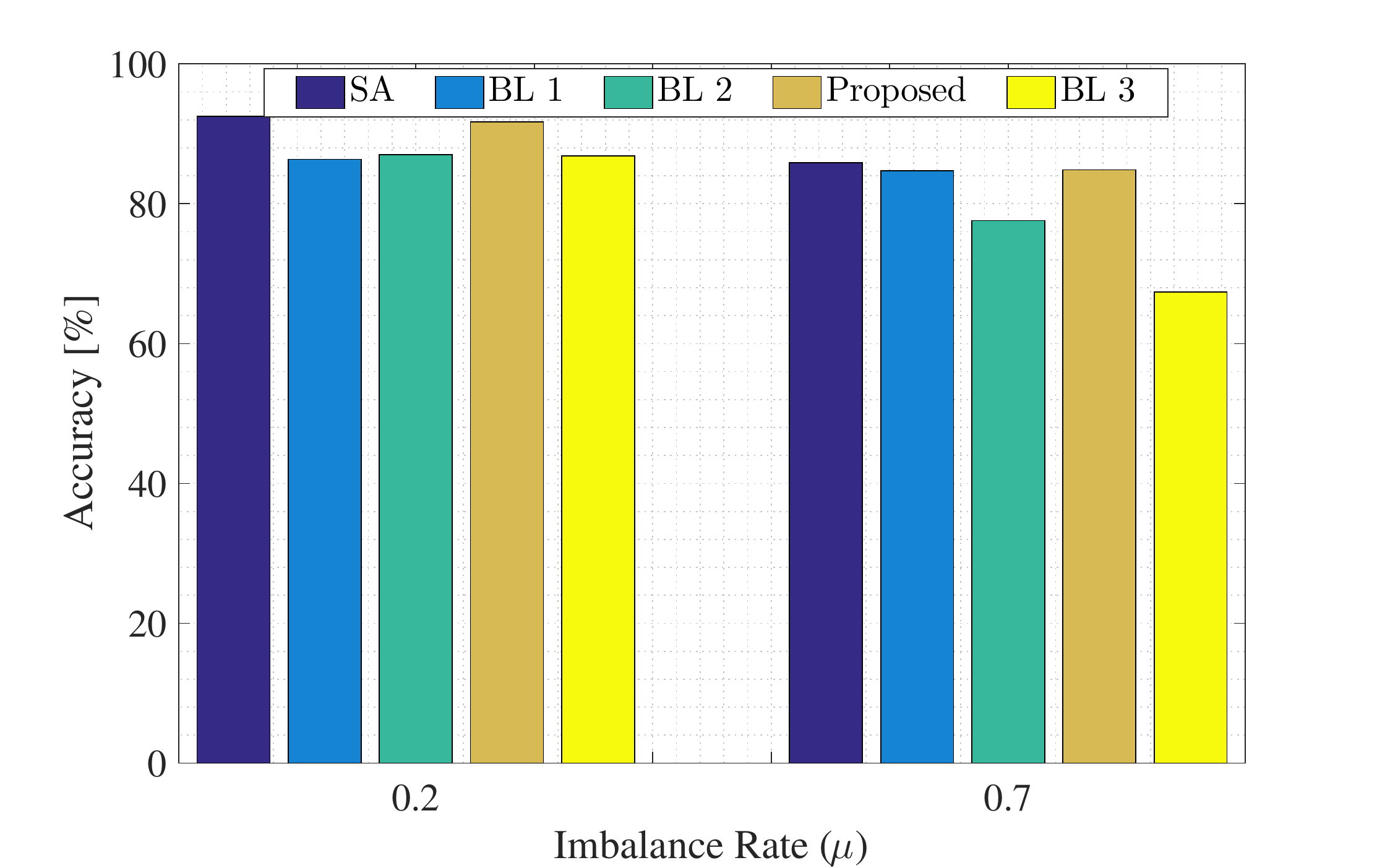}
     \caption{Testing accuracy comparison for $\mu=0.2,0.7$. }
     \label{fig:f5}
   \end{figure}

\begin{figure}[htbp]
     \centering
     \includegraphics[height=1.9in,width=3.1in]{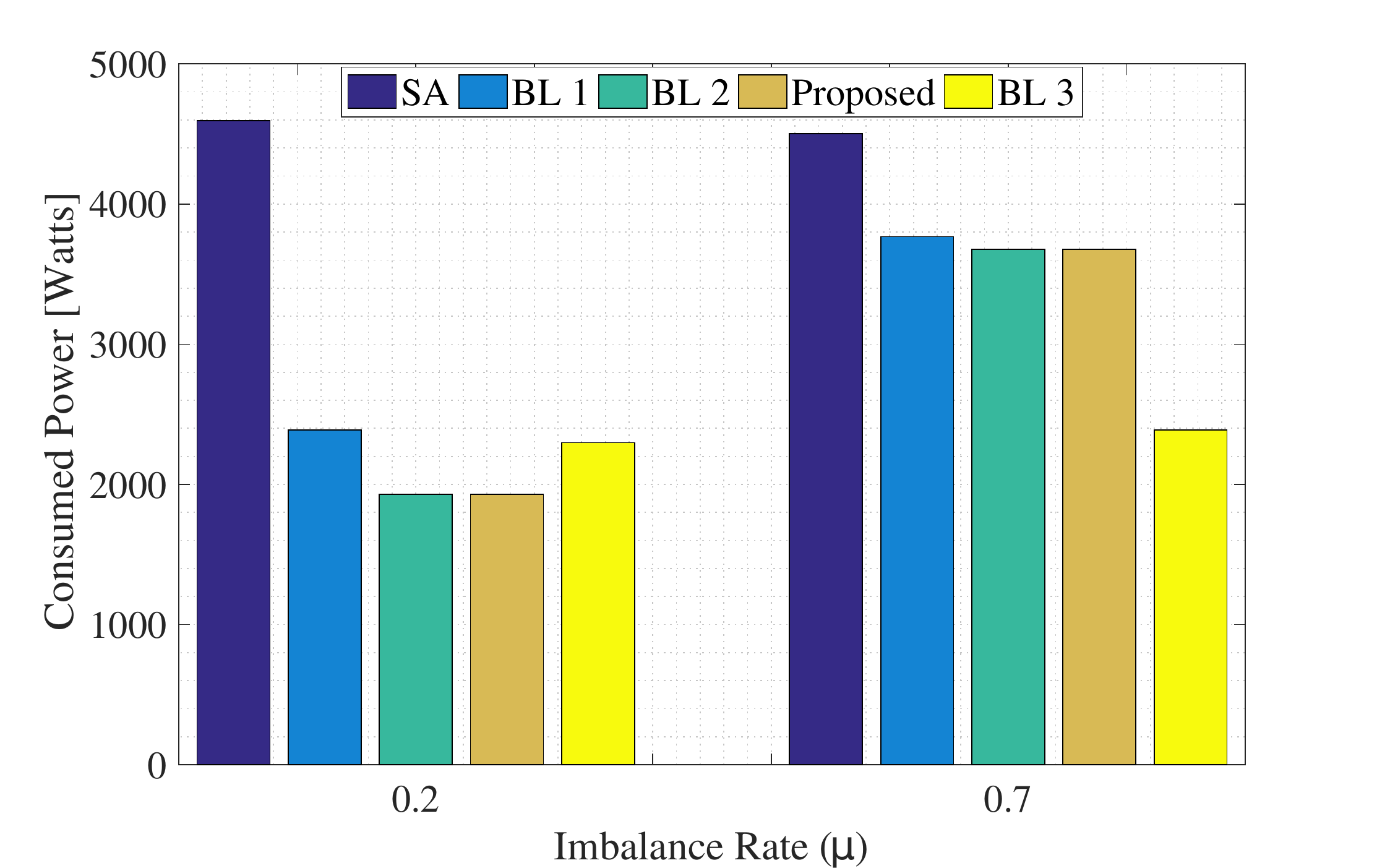}
     \caption{Consumed power comparison for $\mu=0.2,0.7$.}
     \label{fig:f6}
   \end{figure}

Figs. \ref{fig:f5} and \ref{fig:f6} compare the performance of our proposed algorithm with four baseline algorithms, namely a Select All (SA) algorithm, Baseline 1 (BL 1), Baseline 2 (BL 2) and Baseline 3 (BL 3). In particular, the SA algorithm selects all available UAVs, i.e. mimicking the performance of the standard \textsc{FedAvg} approach. Furthermore, BL 1 insures that constraint (\ref{con2}) is satisfied while disregarding the power consumption aspect. On the other hand, BL 2 and 3 are aimed to address the minimization of the power consumption objective, while neglecting constraint (\ref{con2}). However, BL 2 is designed to select the same number of users selected using the proposed algorithm, while BL 3 could select lower number of users satisfying the power consumption objective. 

The results show that our proposed solution outperforms the baseline algorithms. For instance, while selecting all UAVs (SA) yields high testing accuracy results, it results in a higher power consumption figure than the rest of the baselines. Moreover, BL 3 yields a lower power consumption level when compared with the rest of the baselines for $\mu=0.7$, but results in lower classification accuracy (around $67\%$), which is significantly lower than the one resulting form our proposed algorithm (around $83\%$). Also, due to the constraint in (\ref{con3}), our proposed solution shows an improvement in the testing accuracy when compared with BL 1 and BL 2.


\section{Conclusion}
In this paper, we studied the problem of enhancing the accuracy of a learning problem using UAVs. Specifically, we investigated an FL setup to reach high testing accuracy levels in a UAV swarm while maximizing the availability of the UAVs, all while conducting a task such as object recognition. The accuracy was enhanced by constraining the selected UAVs with the aim of resolving the problem of class imbalance in the considered FL setup. Our algorithm showed significant improvements in terms of accuracy, power consumption and availability when compared with several baselines. 

\section{Acknowledgement}

This paper was made possible by PDRA grant \#5-0424-19005 from the Qatar National Research Fund (a member of Qatar Foundation) and the Qatar University Internal Grant IRCC-2020-001. The statements made herein are solely the responsibility of the authors.
\bibliographystyle{ieeetr}
\bibliography{sec_00_main}

\end{document}